\documentclass[letterpaper, 10 pt, conference]{ieeeconf}  

\IEEEoverridecommandlockouts                              

\usepackage{amsmath} 
\usepackage{amsfonts}
\usepackage{tabularx}
\usepackage{booktabs}
\usepackage{graphicx}
\usepackage{url}
\usepackage{hyperref}
\usepackage{algpseudocode}
\usepackage{pgfplots}
\pgfplotsset{compat=1.18}
\usepgfplotslibrary{groupplots}
\usepackage{xstring}
\usepackage[inkscapeformat=pdf]{svg}
\usepackage{svg}
\usepackage{adjustbox}
\usepackage[capitalise]{cleveref}
\usepackage{subcaption}

\title{\LARGE \bf Contrast \& Compress: Learning Lightweight Embeddings for Short Trajectories}

\author{Abhishek Vivekanandan$^{1, 2}$, Christian Hubschneider$^{1, 2}$ and J. Marius Zöllner$^{1,2}$
\thanks{$^{1}$ FZI Research Center for Information Technology, 76131 Karlsruhe, Germany.
        {\tt\small \{vivekana, hubschneider, zoellner\}@fzi.de }}%
\thanks{$^{2}$ Karlsruhe Institute of Technology (KIT), Germany.}%
}
\begin{document}

\maketitle
\thispagestyle{empty}
\pagestyle{empty}

\begin{abstract}
The ability to retrieve semantically and directionally similar short-range trajectories with both accuracy and efficiency is foundational for downstream applications such as motion forecasting and autonomous navigation. However, prevailing approaches often depend on computationally intensive heuristics or latent anchor representations that lack interpretability and controllability. In this work, we propose a novel framework for learning fixed-dimensional embeddings for short trajectories by leveraging a Transformer encoder trained with a contrastive triplet loss that emphasize the importance of discriminative feature spaces for trajectory data. We analyze the influence of Cosine and FFT-based similarity metrics within the contrastive learning paradigm, with a focus on capturing the nuanced directional intent that characterizes short-term maneuvers. Our empirical evaluation on the Argoverse 2 dataset demonstrates that embeddings shaped by Cosine similarity objectives yield superior clustering of trajectories by both semantic and directional attributes, outperforming FFT-based baselines in retrieval tasks. Notably, we show that compact Transformer architectures, even with low-dimensional embeddings (e.g., 16 dimensions, but qualitatively down to 4), achieve a compelling balance between retrieval performance (minADE, minFDE) and computational overhead, aligning with the growing demand for scalable and interpretable motion priors in real-time systems. The resulting embeddings provide a compact, semantically meaningful, and efficient representation of trajectory data, offering a robust alternative to heuristic similarity measures and paving the way for more transparent and controllable motion forecasting pipelines.
\end{abstract}


\section{Introduction}
\label{sec: introduction_related_works}

Accurate trajectory candidate retrieval for motion forecasting remains relatively underexplored, particularly with pre-trained embeddings that can replace heuristic anchors and enable more flexible, scene-compliant prior injections through directional anchors.  
The core task is to accurately and efficiently retrieve trajectories from a trajectory database for a given query trajectory, a fundamental operation for downstream tasks like motion forecasting, where current work focuses on learning this implicitly using latent query anchors \cite{nayakanti_wayformer_2022, zhou_query-centric_2023}. We diverge from implicit learning and explicitly pre-train an embedding model, which could functions as a controllable and interpretable anchor \cite{tas2025wordsmotionextractinginterpretable} generator.
\setlength{\abovecaptionskip}{2pt}
\begin{figure}[t]
\centering
\includegraphics[width=0.90\columnwidth]{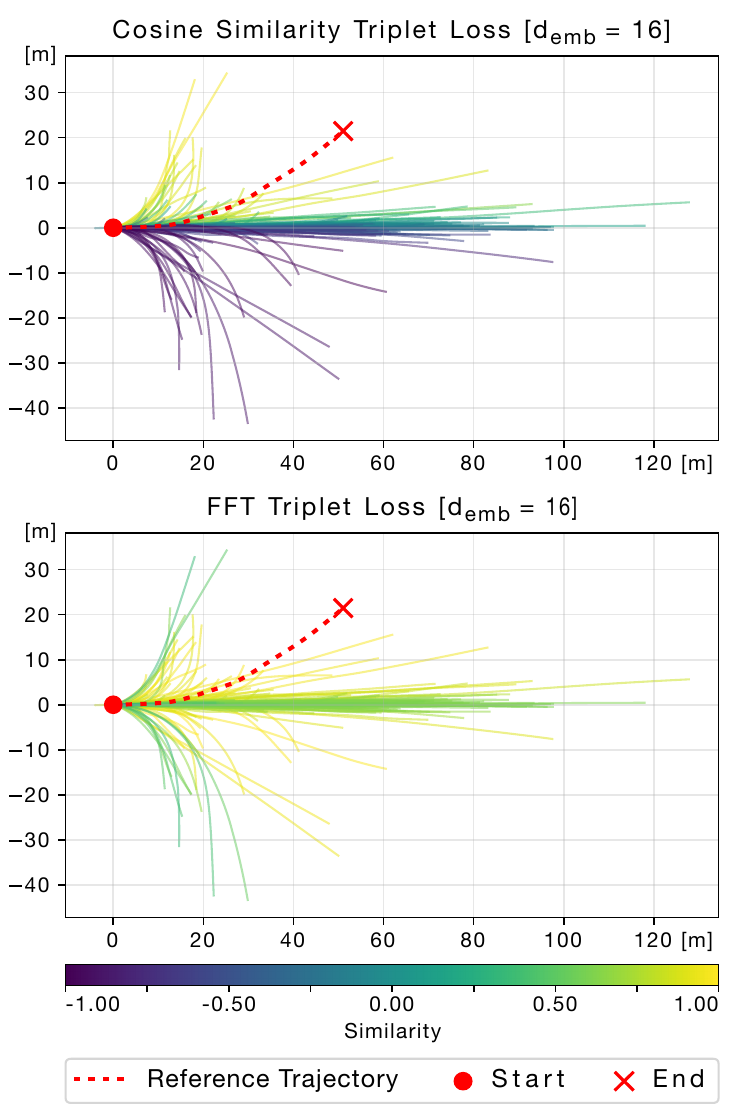}
\caption{Comparison of retrieved trajectories using 16-dimensional embeddings ($d_{emb}=16$) trained with Cosine Similarity (Top) vs. FFT Triplet Loss (Bottom). Trajectories are colored by their similarity score (calculated using the respective metric) to the reference trajectory (red dotted line). Axes are in spatial trajectory coordinates.
Note the superior directional clustering achieved with Cosine similarity.}
\label{fig:cosine_vs_+fft}
\end{figure}

Since this work mainly addresses the motion forecasting task, the ability to retrieve relevant trajectories in spatial space for a given query enables us to move beyond heuristic anchor selection methods — such as KNNs used in prior works \cite{shi_motion_2023, sun_controlmtr_2024} by leveraging embedding-based queries of real, recorded trajectories that better align with road topology and driving intentions.

Traditional approaches, including pairwise distance calculations using DTW, Fréchet, or Hausdorff metrics, are not only computationally expensive but also limited in their ability to capture the nuanced semantics and complex spatio-temporal dynamics of short-range, noisy trajectories. These methods rely on simple geometric or pointwise comparisons, making them sensitive to noise and discretization errors, especially in scenarios where subtle directional changes are critical for inferring intent. Furthermore, heuristic distance measures, such as DTW, do not satisfy the triangle inequality, fundamentally restricting the ability to represent these distances accurately within a Euclidean embedding space making it hard to fine-tune the embedding model with your downstream task.
Alternative strategies that utilize kinematic models or compressed trajectory sets \cite{biktairov_prank_2020, phan-minh_covernet_2020, vivekanandan_ki-pmf_2024, vivekanandan_efficient_2024} may ensure physical plausibility, but often fail to capture the diversity, learned interaction patterns, and subtle intentions present in real-world data.

These limitations — both in computational efficiency and representational expressiveness — motivate the adoption of learned trajectory embeddings. The core idea is to create a model, such as a Transformer, to map variable-length trajectories into fixed-dimensional latent vectors (embeddings) that encapsulate rich semantic information. This approach offers several key advantages:
\begin{itemize}
    \item \textbf{Computational Efficiency:} Once trajectories are embedded into low-dimensional vectors (which can be precomputed), similarity queries reduce to efficient distance computations (e.g., Euclidean, Manhattan, Cosine) between embeddings. This enables the use of scalable search algorithms like FAISS \cite{johnson2019billion}, dramatically reducing query times compared to the $O(N^2)$ complexity of heuristic methods, particularly for large-scale or batched queries (cf. \cite{chang_trajectory_2024}).
    \item \textbf{Enhanced Representation Power:} Learned embeddings can capture complex, high-level features such as velocity profiles, temporal dependencies, directional intent, and even contextual cues from the surrounding scene or agent type. By optimizing with objectives like contrastive learning, embeddings can be explicitly trained to group trajectories with similar underlying intentions or behaviors-even when their raw point sequences differ significantly. 
    \item \textbf{Latent Compression for Downstream Tasks:} Pre-trained embeddings serve as compact latent representations that downstream motion forecasting models can query during learning. This allows for direct use of these embeddings as directional priors or anchors, effectively encoding intent or goals in a manner that is both efficient and semantically meaningful.
\end{itemize}

\section{Related Work}
\label{sec:related_works}
The task of measuring trajectory similarity is well-established, with foundational work relying on heuristic metrics. Notable examples include Dynamic Time Warping (DTW) \cite{salvador_fastdtw_nodate}, Hausdorff distance \cite{huttenlocher_comparing_1993}, and Edit Distance on Real sequences (ERP) \cite{10.5555/1316689.1316758}. While interpretable, these methods often struggle with the specific demands of short-range motion analysis, particularly in robustly capturing directional nuances amidst sensor noise and discretization challenges inherent in fine-grained movements \cite{chang_contrastive_2023, liang_survey_2024}. 
Learned trajectory embeddings have emerged as a powerful alternative, leveraging neural architectures such as Transformers \cite{vaswani_attention_2017} to create fixed-dimensional representations from variable-length sequences. This approach has shown promise for both efficiency, especially using indexing structures like FAISS for fast retrieval \cite{johnson_billion-scale_2019}, and capturing complex spatio-temporal features. However, a significant portion of research utilizing trajectory embeddings has focused on long-range applications, such as analyzing GPS tracks spanning kilometers \cite{hu_spatio-temporal_2023}. For instance, algorithms like TRACLUS \cite{10.1145/1247480.1247546} excel at identifying common sub-trajectories within large datasets, targeting coarser spatial patterns suitable for tasks like route recommendation or broad clustering. An intermediate work which combines the heuristics and learning paradigm can be seen from the works of \cite{deng_learning_2024} where they use a hash based schema with a learnable encoder for trajectory data to perform top-$k$ search on the database.
This prevalent focus on long-range analysis \cite{trajGAT, deng2022efficient, 8731427}  differs fundamentally from the requirements of short-range trajectory retrieval pertinent to motion forecasting. In safety-critical scenarios like intersection navigation, the fine-grained details of directionality and subtle motion changes, which encode agent intent, become crucial. While recent studies explore embeddings for trajectory similarity computation \cite{chang_trajectory_2024} or directly learning the embeddings for a downstream task \cite{halawa_action-based_2022}, our work distinguishes itself by concentrating explicitly on the retrieval and the generated embeddings quality for short-range trajectories. We specifically investigate how contrastive learning, tailored with cosine similarity, can produce embeddings that are highly sensitive to the directional information critical for these short-range maneuvers, aiming to provide effective priors for downstream motion forecasting tasks. 
\section{Methodology}
\label{sec:methodology}
Our approach centers on learning fixed-dimensional vector representations (embeddings) for variable-length trajectories using a contrastive learning framework. The core idea is to train a model that maps trajectories into a latent space where the distance between embeddings reflects the semantic similarity of the original trajectories, particularly capturing nuances crucial for short-range motion forecasting like directional intent. We employ a Transformer-based encoder architecture for this task, to not only leverage its strength in processing sequential data but to also make sure it fits with the downstream architectures which usually employs attention layers.

\subsection{Input Representation and Preprocessing}
\label{subsec:input_rep}
Input trajectories, represented as sequences of $(x, y)$ coordinates over time, undergo a preprocessing step to ensure consistency and focus on the intrinsic motion pattern. As described in our experimental setup (Section V) and following the work \cite{vivekanandan_efficient_2024}, each trajectory is normalized by:
\begin{enumerate}
    \item \textbf{Translation:} Shifting the trajectory so that its starting point is at the origin (0,0).
    \item \textbf{Rotation:} Rotating the trajectory so that it generally aligns with a canonical direction (e.g., facing east / along the x-axis).
\end{enumerate}
This preparation helps the model learn shape and directional features independent of the absolute starting position and orientation in the original scene, as is the case with agent centric motion forecasting.

Input coordinate sequences (60x2 flattened to 120x1) are padded to a fixed 1024-dimension vector (chosen to accommodate typical Transformer input sizes and potential future extensions) and padded sequences are masked during attention. Positional encodings are added to this vector to capture the crucial temporal dynamics of the trajectory.

\subsection{Model Architecture: Transformer Encoder}
\label{sec:model_arch}
At the heart of our methodology lies a simple Transformer-based encoder architecture \cite{vaswani_attention_2017}, chosen for its proven efficacy in capturing long-range sequential dependencies\cite{zhou_informer_2021}. The encoder is composed of $L$ identical $"Layers"$ stacked sequentially, with each layer containing $H$ parallel attention $"Heads"$. Unlike \cite{chang_contrastive_2023} where the authors delved into architectural focus on fusing the spatio-temporal aspects, we strongly believe and also justify through the results as discussed in \cref{sec:experiments} that parameter-light encoder models perform relatively well given proper learning methods.

Given a preprocessed input sequence $\mathbf{x} = x_1 \cdots x_N$, we first project the input tokens into a continuous vector space using an input projection layer, forming the initial embedding matrix $\mathbf{X}^0 = [x_1, \cdots, x_N] \in \mathbb{R}^{N \times d_{\text{model}}}$, where $d_{\text{model}}$ denotes the model's hidden dimension. These embeddings are then iteratively refined through the stack of $L$ encoder layers:

\begin{equation}
\mathbf{X}^l = \text{EncoderLayer}(\mathbf{X}^{l-1}), \quad l \in [1, L].
\end{equation}

Each $\text{EncoderLayer}(\cdot)$ incorporates multi-head self-attention mechanisms, allowing the model to weigh the importance of different points in the trajectory when representing any given point, followed by feed-forward networks producing the desired embedding vector $d_{emb}$. We utilize an efficient attention implementation, such as \textit{FlashAttention} \cite{dao_flashattention_2022}, to manage computational costs.

To mitigate overfitting and enhance generalization, we strategically employ dropout regularization. Dropout layers are introduced primarily after the input projection layer ($p = 0.3$) and also within the attention layers themselves ($p = 0.2$). This injects noise during training, forcing the model to learn more robust representations and simulating real-world scenarios where observations might be partially missing or noisy (e.g., due to sensor occlusions). 

\subsection{Contrastive Learning Framework}
\label{subsec:constrastive_learning}
We train the Transformer encoder using a contrastive learning objective, specifically using the Triplet Loss \cite{Schroff2015FaceNetAU}, to structure the embedding space meaningfully.

\subsubsection{Similarity Metrics for Triplet Selection}
A crucial step in contrastive learning is defining what constitutes “similar” (positive) and “dissimilar” (negative) pairs relative to an anchor trajectory. We explored two primary similarity metrics computed directly on the input trajectory data:

\begin{itemize}
    \item \textbf{Cosine Similarity} We first compute pairwise trajectory distances \(d_{ij}\) (e.g., Average Displacement Error, ADE) between sequences. Then, we extract each trajectory’s overall displacement vector \(\Delta \mathbf{p}_i = \mathbf{p}_i^{\rm end} - \mathbf{p}_i^{\rm start}\). The cosine similarity gives the directional affinity \(\cos(\Delta \mathbf{p}_i, \Delta \mathbf{p}_j)\). We combine these by weighting the directional similarity by the inverse distance via the formula:
    \begin{equation}
    s_{ij} = \frac{1}{1 + \alpha \,d_{ij}}\;\cos\bigl(\Delta \mathbf{p}_i,\Delta \mathbf{p}_j\bigr)    
    \end{equation}
    where \(\alpha\) is a weighting factor (kept at 0.5 consistently) balancing spatial proximity and directional alignment.

    \item \textbf{FFT similarity} Each trajectory sequence is transformed along its time axis via the Fast Fourier Transform (FFT), with the intention of learning a representation of a trajectory's general shape. We retain the magnitude of the first \(\lfloor T/2\rfloor+1\) frequency coefficients while discarding phases which are less relevant for capturing the trajectory's  shape (where \(T\) is the sequence length), also leveraging the conjugate symmetry for real inputs. These magnitudes are flattened into a feature vector for each trajectory, which is then \(L_2\)-normalized to ensure scale invariance. The pairwise spectral similarity matrix is efficiently computed as the matrix product of these normalized feature vectors with their transpose (equivalent to cosine similarity).
\end{itemize}

The described similarity metrics (Cosine or FFT) define positive and negative examples for our contrastive training process. Each training instance comprises an anchor, a positive (a trajectory within the batch deemed similar to the anchor above a threshold), and a negative. For selecting the negative example, we utilize Random Mining. This strategy contrasts with hard-negative mining approaches that specifically target negatives closest to the anchor in the embedding space. Instead, random mining selects any trajectory from the batch that is dissimilar to the anchor (below the similarity threshold) with uniform probability. This computationally simpler method ensures broad coverage of the negative distribution, enhances training stability, and is implemented by sampling one random negative for each anchor-positive pair. Large batch sizes naturally complement random mining by increasing the likelihood of sampling informative negatives, a principle reflected in our experimental design (\cref{subsec:training_methodology}).

\subsubsection{Triplet Loss Function}
\label{sec:triplet_loss}

We train the trajectory encoder using a triplet loss objective designed to learn discriminative embeddings. For each training triplet, consisting of the anchor (\(a\)), a positive (\(p\)), and a negative (\(n\)) trajectory, the encoder produces corresponding output embeddings. These embeddings are first \(L_2\)-normalized, projecting them onto the unit hypersphere. This normalization improves training stability and ensures that the loss focuses on the angular separation between embeddings.

The triplet loss \(\mathcal{L}_{\text{triplet}}\) is then calculated based on the distances between these normalized embeddings:
\begin{equation}
\label{eq:triplet_loss_formula} 
\mathcal{L}_{\text{triplet}} = \max\bigl(0, d(e_a, e_p) - d(e_a, e_n) + m\bigr)
\end{equation}
Here, \(e_a, e_p, e_n\) represent the normalized embeddings for the anchor, positive, and negative samples, respectively. \(d(\cdot, \cdot)\) denotes a distance metric in the embedding space (e.g., Euclidean distance), and \(m > 0\) is a predefined margin hyperparameter. Minimizing this loss encourages the distance between the anchor and positive (\(d(e_a, e_p)\)) to be smaller than the distance between the anchor and negative (\(d(e_a, e_n)\)) by at least the margin \(m\).

\subsection{Final Embedding Generation}

This resulting vector captures an average representation across the entire trajectory's hidden states. 

We denote the transformer’s output as  
\[
H\in\mathbb{R}^{B\times T\times d_{\mathrm{model}}},
\]  
where B = batch size, T = sequence (temporal) length, \(d_{\mathrm{model}}\) = hidden (feature) dimension which is maintained at $512$ throughout the model.  

We then apply temporal average pooling to collapse the time axis:  
\[
c \;=\;\frac{1}{T}\sum_{t=1}^{T}H_{:,\,t,:\;} 
\;\in\;\mathbb{R}^{B\times d_{\mathrm{model}}}
\]  
The resulting context vector \(c\) or $d_{emb}$ is a fixed-length context embedding that captures the mean representation of the entire trajectory’s hidden states.

\section{Evaluation Metrics}

The evaluation of trajectory generation or retrieval systems often involves comparing a ground truth query trajectory, denoted as $q$, against a set of $K$ candidate trajectories, $\{T'_1, \dots, T'_K\}$, generated or retrieved by the model.
The \textbf{minADE} metric for this query is defined as the minimum Average Displacement Error (ADE) computed between the query $q$ and each of the $K$ candidates. The displacement metrics reported throughout this works are in $meters\ (m) $ and $K=6$ unless stated otherwise:
\[
\text{minADE}(q, \{T'_k\}_{k=1}^K) = \min_{1 \le k \le K} \text{ADE}(q, T'_k)
\]
Analogously, \textbf{minFDE} represents the minimum Final Displacement Error (FDE), focusing solely on the final end point of a trajectory:
\[
\text{minFDE}(q, \{T'_k\}_{k=1}^K) = \min_{1 \le k \le K} \text{FDE}(q, T'_k)
\]
Conversely, the \textbf{avgADE} metric calculates the average ADE across the $K$ candidates:
\[
\text{avgADE}(q, \{T'_k\}_{k=1}^K) = \frac{1}{K} \sum_{k=1}^{K} \text{ADE}(q, T'_k)
\]
Similarly, \textbf{avgFDE} computes the average FDE:
\[
\text{avgFDE}(q, \{T'_k\}_{k=1}^K) = \frac{1}{K} \sum_{k=1}^{K} \text{FDE}(q, T'_k)
\]
These per-query metrics are typically aggregated, by averaging, over an entire dataset of query set $\mathcal{Q}$ to report overall similarity performance in the tables in ~\cref{sec:experiments}.
\section{Experimental setup}
\label{sec:experimental_setup}
\subsection{Dataset and Preprocessing}
We use the Argoverse 2 Motion Forecasting Dataset~\cite{wilson_supplemental_nodate}, which includes diverse traffic scenarios. For each scenario, the focal agent's future trajectory is extracted and considered as a training sample. Following the agent-centric normalization described in ~\cref{subsec:input_rep}, trajectories are translated to the origin and rotated to a canonical direction, yielding a standardized bank for training~\cite{biktairov_prank_2020, vivekanandan_ki-pmf_2024}.

The dataset is split: 50\% of the validation set is used for monitoring, and the remaining 50\% serves as a hold-out test set. All training data is used in full. Each trajectory is represented as a sequence of 60 $(x, y)$ coordinates; sequences are masked or padded to fit the Transformer's fixed input size (1024 dimensions, see ~\cref{sec:model_arch}). 


\subsection{Training Methodology}
\label{subsec:training_methodology}
The Transformer encoder (~\cref{sec:methodology}) is trained with a contrastive Triplet Loss (Eq.~\ref{sec:triplet_loss}) for 4900 (cosine similarity) and 19400 (FFT similarity) steps, corresponding to 100 epochs on 197k samples. We use OneCycleLR~\cite{cycleLR} with PyTorch v2.4.0~\cite{paszke2019pytorchimperativestylehighperformance} defaults. Online batch mining forms triplets (anchor, positive, negative) within each batch, based on input-space similarity: positives have a score $\geq 0.7$ and negatives $<0.7$. Output embeddings are normalized to the unit sphere during the \textit{triplet loss mining}. Triplets are randomly selected from the respective candidate pools following the \textbf{random mining} approach (cf. Section~\ref{subsec:mining_strategy}). Large batch sizes (up to 1024 for FFT, 4096 for Cosine) are enabled by \texttt{bf16} mixed-precision training~\cite{micikevicius_mixed_2018} on NVIDIA A100 GPUs, which also improves test accuracy by up to 4 percentage points over \texttt{fp16}~\cite{choi_learning_2020}. FFT-based mining requires more memory due to storage of coefficient magnitudes. The number of triplets per batch is kept proportional across experiments and similarity functions.

\cref{table:cosine_vs_fft_comparison} presents the stark difference between Cosine and FFT similarity for triplet selection (4 Heads, 2 Layers — 4H2L); evaluating the performance on minADE and minFDE on the AV2 test set, metrics commonly used in trajectory forecasting retrieval tasks. This confirms the hypothesis that Cosine similarity, by directly balancing the pairwise and directional alignment, is better suited for capturing the intent of the short-range maneuvers compared to the frequency-focused FFT approach. 

\begin{table}[ht]
\centering
\caption{Performance comparison: Cosine vs. FFT for using Transformer encoder with 4H2L}
\begin{tabularx}{\columnwidth}{cc|cc}
\toprule
\textbf{Similarity Function} & \textbf{$d_{emb}$} & \textbf{minADE} $(\downarrow)$ & \textbf{minFDE} $(\downarrow)$ \\
\midrule
        & 128   & 0.3191 & 0.5554 \\ 
        & 64  & 0.3234 & 0.5131 \\
Cosine  & \textbf{32}  & 0.3271 & \textbf{0.4801} \\
        & \textbf{16}  & \textbf{0.3170} & 0.4869\\ 
        & 8   & 0.3271 & 0.5995 \\
        & 4   & 0.4206 & 0.7500 \\
\midrule 
        & 128 & 1.0074 & 1.8756 \\
        & 64  & 1.0778 & 1.9970 \\
FFT     & 32  & 0.8774 & 1.5750 \\
        & 16  & 1.1273 & 2.0457 \\
        & 8   & 1.2586 & 2.3409 \\
        & 4   & 1.5006 & 2.9046 \\
\bottomrule
\end{tabularx}
\label{table:cosine_vs_fft_comparison}
\end{table}

\begin{figure*}[ht]
    \centering
\includegraphics[width=0.95\textwidth]{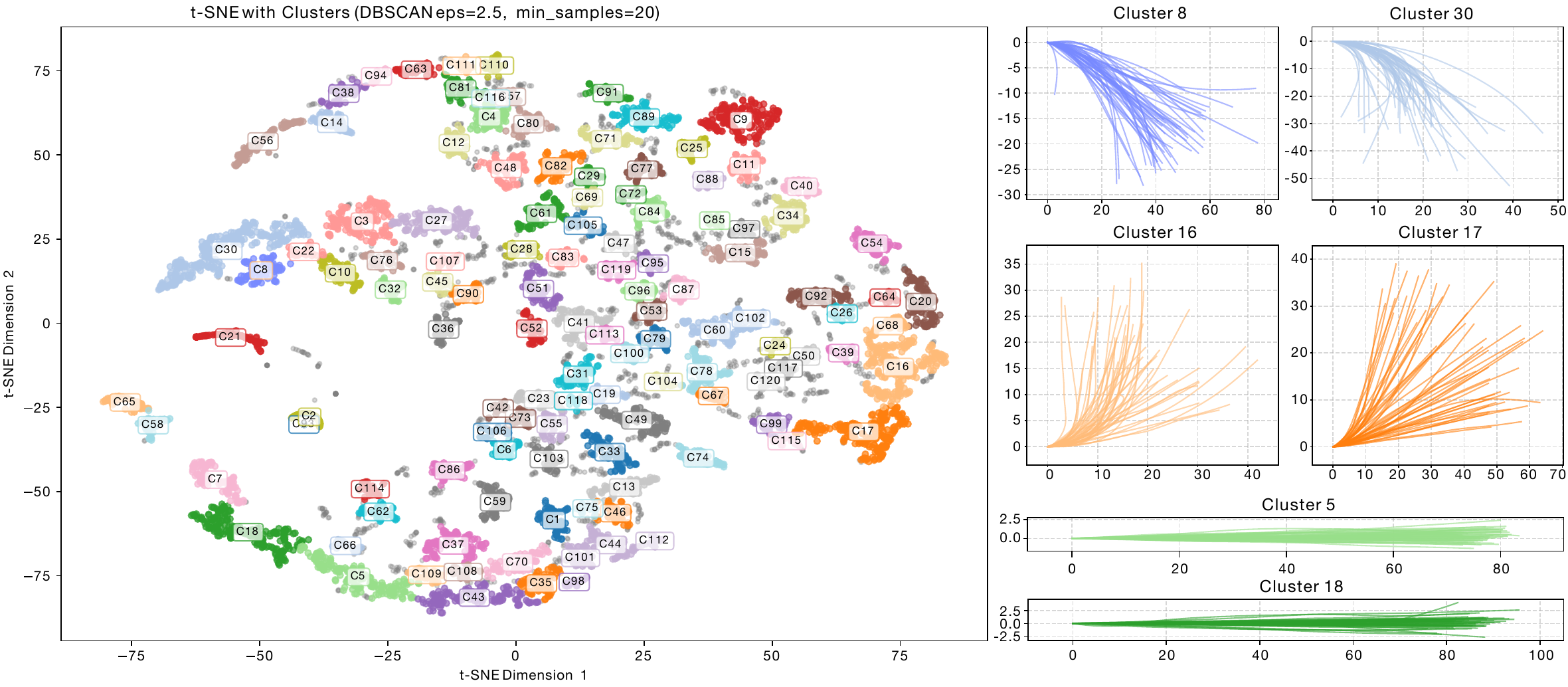}
    \caption{t-SNE visualization demonstrating the structure of the learned trajectory embedding space (4H2L/$d_{emb} = 16$ Cosine model). \textit{(Left)} Embeddings clustered using DBSCAN, where proximity indicates similarity. \textit{(Right)} Corresponding original spatial trajectories for selected clusters, showing clear visual coherence within each group (e.g., Cluster 8: left turns, Cluster 18: straight). This qualitatively confirms that the embeddings capture semantic trajectory similarity. Note the difference in axis scaling.}
    \label{fig:t-sne-cluster}
\end{figure*}

\section{Evaluation}
\label{sec:experiments}

This section evaluates the effectiveness of the learned embeddings. Our primary objectives are to:
\begin{itemize}
    \item Assess whether the learned embeddings create a latent space that supports meaningful cluster based on semantic similarity.
    \item Determine the optimal hyperparameters (network architecture, embedding dimension, similarity metric) for achieving high-quality embeddings.
    \item Evaluate the quality of the resulting embeddings, both quantitatively and qualitatively.
\end{itemize}

\renewcommand{\arraystretch}{0.8} 
\begin{table*}[ht]
\centering
\caption{Performance comparison of different network architectures and respective output embedding sizes trained using Cosine Similarity Triplet Loss. The reported values are averaged across the test dataset} 
\begin{tabularx}{1.3\columnwidth}{cc|cccc}
\toprule
\textbf{Architecture} & \textbf{$d_{emb}$} & \textbf{avgADE} $(\downarrow)$ & \textbf{minADE} $(\downarrow)$ & \textbf{avgFDE} $(\downarrow)$ & \textbf{minFDE} $(\downarrow)$ \\
\midrule
              & 128 & 0.6041 & 0.3191          & 1.4470 & 0.5606 \\
              & 64  & 0.7574 & 0.3506          & 1.8690 & 0.6656 \\
    8H4L      & 32  & 0.7363 & 0.3720          & 1.8484 & 0.6881 \\
              & 16  & 0.6901 & 0.3542          & 1.6991 & 0.6551 \\
              & 8   & 1.0122 & 0.4626          & 2.3935 & 0.8695 \\
              & 4   & 1.0583 & 0.4767          & 2.4849 & 0.9243 \\
\midrule 
              & 128 & 0.7132 & 0.3520          & 1.5169 & 0.5554 \\
              & 64  & 0.5932 & 0.3234          & 1.3424 & 0.5131 \\
    4H2L      & 32  & 0.6238 & 0.3271          & 1.2359 & 0.4801 \\
              & \textbf{16}  & \textbf{0.5783} & \textbf{0.3170} & 1.2365 & 0.4869 \\
              & 8   & 0.6296 & 0.3271          & 1.6300 & 0.5995 \\
              & 4   & 0.9151 & 0.4206          & 2.0843 & 0.7500 \\
\midrule 
              & 128 & 0.6357 & 0.3284          & 1.2696 & 0.4502 \\
              & 64  & 0.6157 & 0.3296          & 1.3432 & 0.5059 \\
    4H1L      & 32  & 0.5955 & 0.3234 & 1.2357 & 0.4693 \\
              & \textbf{16}  & 0.6049 & 0.3241          & \textbf{1.1570} & \textbf{0.4506} \\
              & 8   & 0.6161 & 0.3259          & 1.2836 & 0.4785 \\
              & 4   & 0.8614 & 0.4100          & 1.7973 & 0.6521 \\
\bottomrule
\end{tabularx}
\label{table:architecture_embedding_comparison}
\end{table*}

\subsection{Mining Strategy Evaluation}
\label{subsec:mining_strategy}
For negative sampling in our contrastive learning setup, we primarily utilize \textbf{random mining}. This approach selects any trajectory from the batch deemed dissimilar (below the similarity threshold) to the anchor with uniform probability. The key advantages are its computational efficiency, simplicity to implement, and inherent training stability, particularly crucial when dealing with large datasets and batch sizes.
To assess if more sophisticated sampling was required, we experimented with a \textit{dynamic mining} strategy incorporating phases of hard ($d(A, N) < d(A, P)$) and semi-hard ($d(A, P) < d(A, N) < d(A, P) + \text{margin}$) negative selection. While theoretically promising for accelerating learning on difficult examples, this approach proved sensitive to hyperparameter tuning and showed potential for training instability and overfitting, where the model focuses too much on outlier negatives.
Ultimately, random mining, especially when coupled with the \textbf{large batch sizes} employed in our experiments, proved highly effective. The large batch size inherently increases the likelihood of sampling a diverse range of informative negatives over time, mitigating concerns about overly focusing on common behaviors (like straight driving). This combination yielded robust convergence and the high-quality, stable embedding results presented, demonstrating its suitability for this task without the need for more complex, potentially unstable mining techniques.

\subsection{Choice of Similarity Measure: Cosine vs. FFT}
\label{subsec:similarity_measure}
As quantitatively demonstrated in \cref{table:cosine_vs_fft_comparison}, embeddings trained using \textbf{Cosine similarity} for triplet selection significantly outperformed those trained using FFT-based similarity, achieving substantially lower minADE and minFDE scores. This aligns with our hypothesis (~\cref{sec: introduction_related_works}) that for short-range trajectory analysis, capturing directional information is paramount. Cosine similarity, measuring the angle between vector representations, directly encodes directional alignment.
Conversely, FFT-based similarity focuses on the frequency domain, emphasizing periodic patterns and overall curvature. While useful for some applications, this focus proved detrimental for distinguishing safety-critical maneuvers with similar shapes but opposite directions. For instance, FFT-based embeddings often struggled to separate clusters representing \textbf{left turns versus right turns} initiated from the same point, as both exhibit curvature but in opposing directions, as can be seen from  \cref{fig:cosine_vs_+fft}. Cosine similarity-based embeddings effectively distinguished these cases due to their inherent directionality components.

Based on the quantitative results (\cref{table:architecture_embedding_comparison}) and qualitative analysis (\cref{table:cosine_vs_fft_comparison}), the models trained with \textbf{Cosine similarity} clearly outperform FFT-based models for short-range trajectory embedding.

\subsection{Embedding Dimension and Architecture Choice}
\label{subsec:results_discussion}

The choice of the overall model architecture (in terms of Transformer heads and layers) and the embedding dimension $d_{emb}$ not only impacts the overall model performance, but given possible downstream applications, is also relevant for the computational performance of the trajectory embeddings. To summarize some key observations from \cref{table:architecture_embedding_comparison}:
\begin{itemize}
    \item Smaller architectures (4H1L, 4H2L) generally outperform the larger 8H4L model, suggesting efficiency in capturing relevant features for this task.
    \item An embedding dimension of $d_{emb} = 16$ consistently yields strong performance across different metrics and architectures, achieving the best minADE (with 4H2L) and the best minFDE (with 4H1L). Larger dimensions do not necessarily improve, and smaller dimensions (8, 4) show performance degradation.
    \item The \textbf{4H2L} model with $d_{emb} = 16$ achieves the best minADE (0.3170), indicating superior accuracy in matching the overall trajectory shape on average.
    \item The \textbf{4H1L} model with $d_{emb} = 16$ achieves the best minFDE (0.4506) and the best avgFDE (1.1570), suggesting it excels at predicting the final endpoint accurately. It is also the most computationally efficient model among the top performers.
\end{itemize}

Comparing the top performers: \textbf{4H2L} with \textbf{16} dimensions has the best minADE (overall shape) whereas \textbf{4H1L} with \textbf{16} dimensions has the best minFDE (endpoint accuracy) and avgFDE, plus fewest parameters.
Given that both models are significantly leaner and perform better than the larger 8H4L model, the choice between 4H2L/16dim and 4H1L/16dim depends on the specific application priority. If precisely matching the overall trajectory shape is critical, 4H2L might be preferred. However, if accurate final endpoint prediction and computational efficiency are paramount (often the case in real-time forecasting), the \textbf{4H1L} with \textbf{$d_{emb}=16$} emerges as the most compelling choice. It achieves top FDE performance with the simplest architecture, while still demonstrating very strong ADE scores.

\begin{figure*}[ht]
  \centering
  \begin{subfigure}[b]{\textwidth}
    \centering
    \includegraphics[width=\columnwidth]{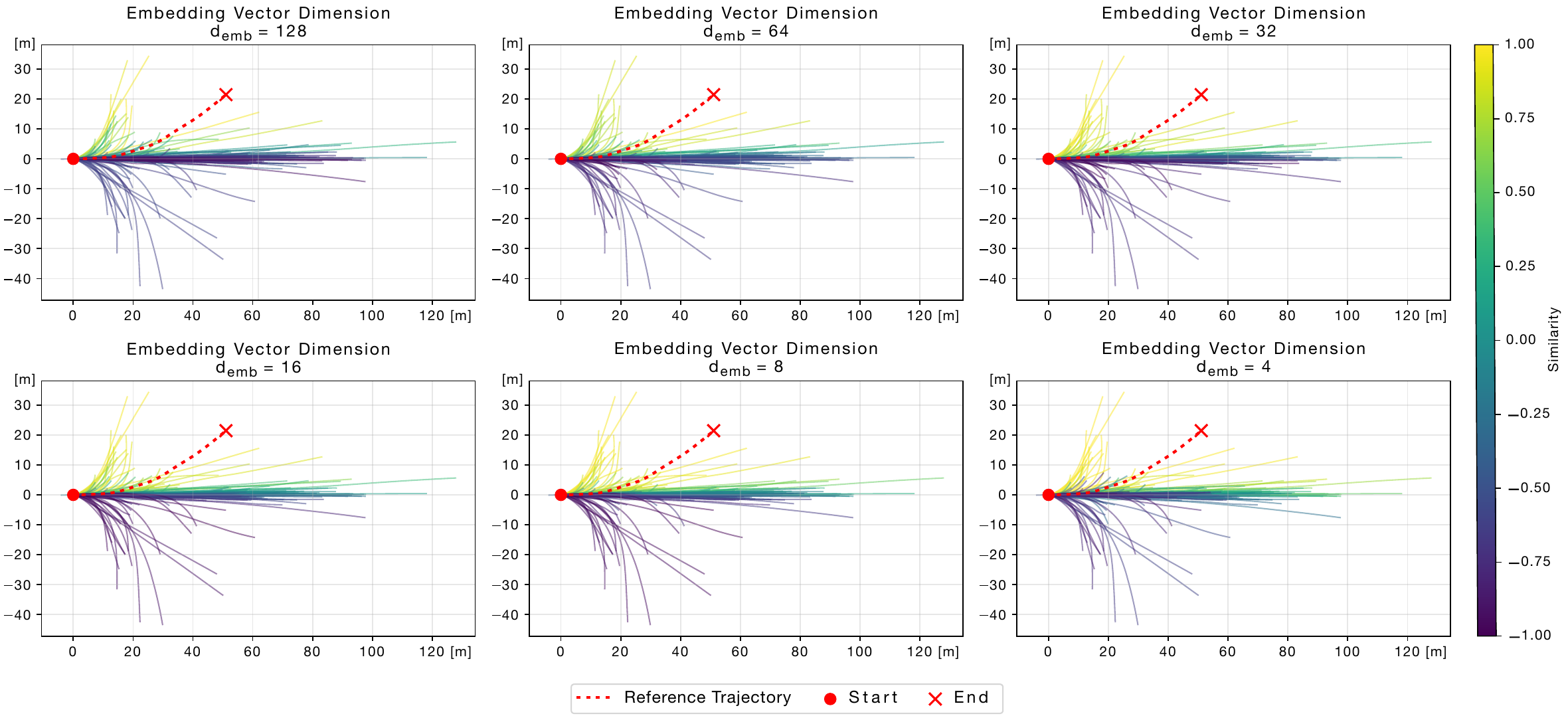}
    \caption{Embedding dimension analysis}
  \end{subfigure}
  \hfill
  \begin{subfigure}[b]{\textwidth}
    \centering
    \includegraphics[width=\columnwidth]{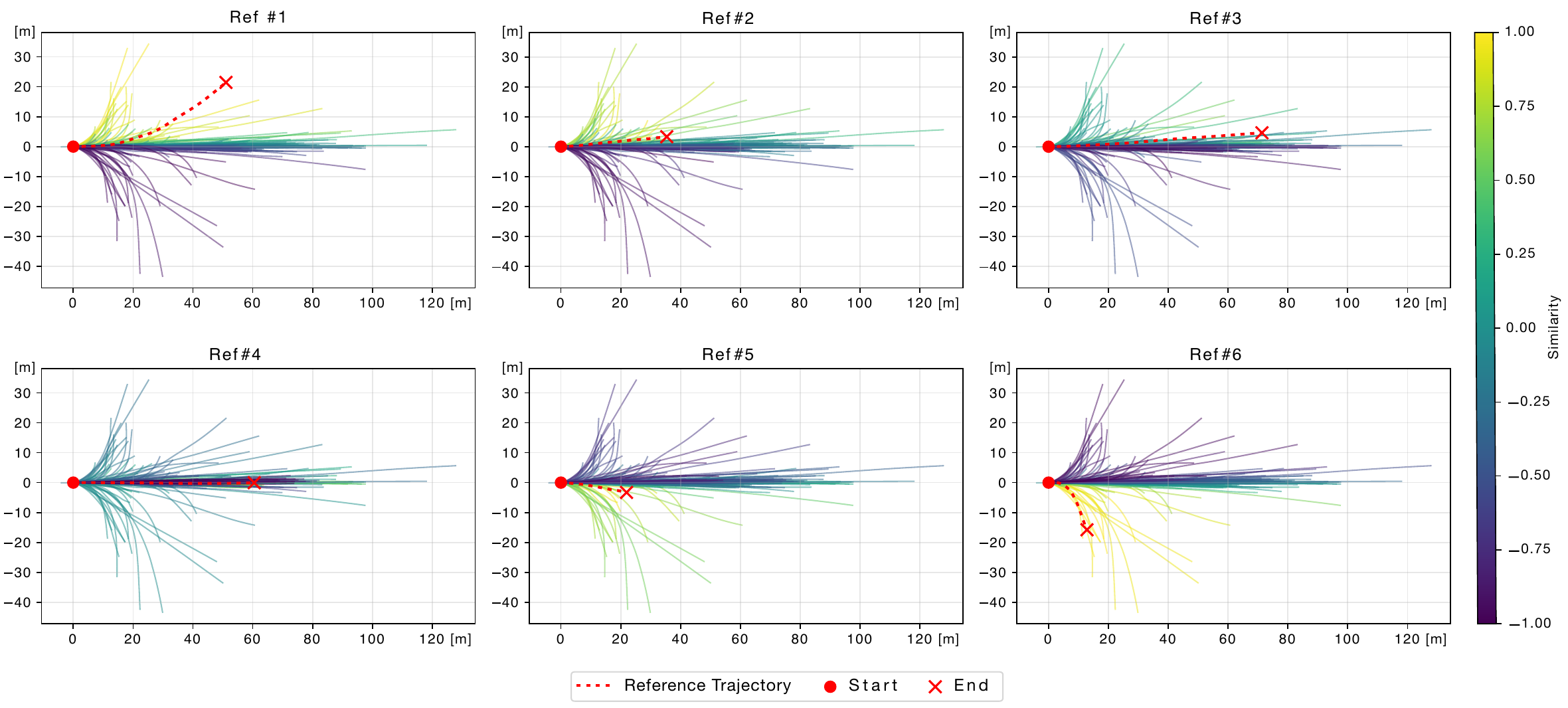}
    \caption{Reference trajectory examples}
  \end{subfigure}
  \caption{a) Trajectory similarity within trajectory sets for different output dimensions based on the trajectory embeddings  from the 4H2L Cosine model. A reference query trajectory (red sampled from test dataset) is shown with its nearest neighbors retrieved from the pre-computed training embeddings. Neighbors are color-coded by their embedding similarity to the query trajectory. Yellow/green indicates high similarity, blue/purple indicates low similarity or dissimilarity. b) Showcases similar visualizations but for different reference trajectories}
  \label{fig:embedding_dim_analysis}
\end{figure*}

In addition to the quantitative results, \cref{fig:embedding_dim_analysis} visualizes the similarity of a random trajectory set for multiple embedding dimensions (a) and for reference trajectories (b). Those results underline that even small embedding dimensions are capable of encoding short trajectories.

\subsection{Qualitative Analysis: Embedding Space Visualization}
\label{subsec:embedding_space}
We qualitatively evaluate the learned embedding space by visualizing trajectory embeddings and their neighbors using t-SNE~\cite{maaten_visualizing_2008}. As shown in~\cref{fig:t-sne-cluster}, the 4H2L Cosine similarity model produces embedding spaces where semantically similar trajectories cluster together, particularly reflecting directional alignment. This correspondence between latent proximity and semantic similarity confirms the effectiveness of the learned representations~\cite{chuang2020debiasedcontrastivelearning}. Notably, even with lower embedding dimensions (16 and 32), the clustering structure remains strong, consistent with quantitative findings.
\section{Exploration of Non-Learned Approaches}
\label{sec:preliminary_investigations}

Before committing to learned trajectory embeddings, we investigated several alternative heuristic and sampling-based methods to assess their feasibility for retrieving or generating relevant short-range trajectories in a real world setting. The goal was to determine if simpler, non-learned approaches could adequately address the requirements of capturing nuanced maneuvers efficiently, without necessitating complex neural network models. Key methods explored included:

\begin{itemize}
    \item Precomputed distance matrix methods.
    \item K-Nearest Neighbor (KNN) retrieval based on trajectory endpoints.
    \item KNN retrieval based on multiple points along a reference path (e.g., lane centerline).
    \item Trajectory generation constrained by vehicle kinematics and lane boundaries.
\end{itemize}

We summarize the findings for the most relevant explorations below.

\subsection{Precomputed Pairwise Distance Matrix}
\label{sec:ade_matrix}
One approach implemented was to precompute all pairwise distances between trajectories in a large reference dataset (e.g., using metrics like ADE, DTW, or Hausdorff) and store these in a matrix. During inference, given a query trajectory, one could compute its distance to all reference trajectories and find the top-$k$ trajectories.

\textbf{Limitations:} While conceptually simple, this method suffers from significant scalability issues.
\begin{itemize}
    \item \textbf{Precomputation Cost:} Calculating all pairwise distances requires $O(N^2)$ computations, where $N$ is the number of trajectories in the reference dataset. This is computationally prohibitive for large datasets.
    \item \textbf{Query Cost:} Finding the nearest neighbor for a new query trajectory still requires computing $N$ distances, resulting in $O(N)$ query time, which can be too slow for real-time applications. For embedding models, we implemented a FAISS-GPU implementation based on IVF for faster-querying.
\end{itemize}

\subsection{KNN Retrieval Based on Trajectory Endpoints}
\label{sec:knn_method1}
This method aims to find similar trajectories based on their final destination. The final $(x, y)$ coordinates of the normalized trajectories were used as features to build a KNN search structure (through a KD-tree). Given a query trajectory, its normalized final point was used to query the KNN structure and retrieve trajectories with the closest endpoints in the normalized space.

\textbf{Limitations:} This approach is effective only when the agent is very close to executing its final maneuver (e.g., already deep into a turn). It fails to capture the agent's intent or differentiate between potential future paths when the agent is further away from the decision point (e.g., approaching an intersection with multiple options), as the endpoint alone provides insufficient context.

\subsection{KNN Retrieval Based on Multiple Points Along a Reference Path}
\label{sec:knn_method2}
To address the limitations of endpoint-based KNN, we explored using multiple query points sampled along a reference path, typically the lane centerline relevant to the query agent. For each sampled point on the centerline, we would query a KNN structure (trained on corresponding points from the normalized reference trajectory dataset) to find trajectories that pass close to that specific intermediate point.

\textbf{Limitations:} While attempting to provide more context than just the endpoint, this method often leads to poor results in practice. Forcing retrieved trajectories to strictly adhere to multiple, specific intermediate query points frequently results in the selection or generation of kinematically implausible or geographically non-compliant trajectories (e.g., paths that go off-road or make unrealistic turns) to satisfy the multiple spatial constraints simultaneously.

\subsection{Summary of Limitations}
These preliminary investigations revealed significant limitations in non-learned approaches for our target application. Heuristic matching methods often suffer from high computational cost and may fail to capture the necessary semantic nuances crucial for short-range maneuvers. Sampling and generation methods, while ensuring kinematic plausibility, struggle with efficiency, diversity, and incorporating learned interaction patterns. These findings motivated our exploration of learned trajectory embeddings using contrastive learning, aiming to create representations that are both computationally efficient for retrieval and semantically rich, capturing the essential characteristics of short-range trajectories.
\section{Conclusion}

This work demonstrates the effectiveness of contrastive learning with Cosine similarity for generating fixed-dimensional embeddings of short trajectories. We show that this approach, using compact Transformer architectures (notably 4H1L) and low-dimensional embeddings ($d_{emb}=16$), yields superior retrieval performance, particularly in capturing directional intent crucial for motion forecasting, compared to FFT-based similarity and larger models. The resulting embeddings offer an efficient, semantically meaningful, and interpretable alternative to heuristic methods, providing robust priors for downstream tasks.

\renewcommand{\arraystretch}{0.8} 
\begin{table*}[ht] 
\centering
\scriptsize 
\caption{Performance comparison for different model dropout rates with fixed input dropout (0.2). Architectures and embedding dimensions varied. Best results per dropout group are bolded.}
\label{tab:model_dropout_templated}
\begin{tabularx}{1.2\columnwidth}{cc|cccc} 
\toprule
\textbf{Architecture} & \textbf{$d_{emb}$} & \textbf{avgADE} $(\downarrow)$ & \textbf{minADE} $(\downarrow)$ & \textbf{avgFDE} $(\downarrow)$ & \textbf{minFDE} $(\downarrow)$ \\ 
\midrule
\multicolumn{6}{@{}l}{\textbf{Model Dropout = 0.001} (Input Dropout = 0.2 fixed)} \\ 
\addlinespace[0.8ex]
              & 128       & 0.6006     & 0.3147     & 1.4726     & 0.5602     \\
              & 64        & 0.6626     & 0.3202     & 1.6029     & 0.5698     \\
    8H4L      & 32        & 0.5765     & 0.3061     & 1.4459     & 0.5481     \\
              & 16        & 0.5619     & 0.3024     & 1.4024     & 0.5404     \\
\midrule 
              & 128       & 0.5585     & 0.3104     & 1.3826     & 0.5345     \\
              & 64        & 0.5609     & 0.3075     & 1.2915     & 0.4975     \\
    4H2L      & 32        & 0.5473     & 0.3062     & 1.2745     & 0.4992     \\
              & 16        & 0.5749     & 0.3088     & 1.2840     & 0.4897     \\
\midrule 
              & 128       & \textbf{0.5160} & \textbf{0.2929} & \textbf{1.0471} & \textbf{0.4072} \\
              & 64        & 0.5691     & 0.3121     & 1.1611     & 0.4520     \\
    4H1L      & 32        & 0.5204     & 0.2949     & 1.1166     & 0.4342     \\
              & 16        & 0.5603     & 0.3051     & 1.1930     & 0.4686     \\
\midrule[\heavyrulewidth] 
\multicolumn{6}{@{}l}{\textbf{Model Dropout = 0.5} (Input Dropout = 0.2 fixed)} \\ 
\addlinespace[0.8ex]
              & 128       & 0.8643     & 0.4188     & 2.0099     & 0.7333     \\
              & 64        & 0.6360     & 0.3363     & 1.4607     & 0.5630     \\
    8H4L      & 32        & 0.9722     & 0.4447     & 2.2405     & 0.8035     \\
              & 16        & 0.9000     & 0.4318     & 1.9705     & 0.7248     \\
\midrule 
              & 128       & 0.6998     & 0.3531     & 1.4926     & 0.5360     \\
              & 64        & 0.8535     & 0.4159     & 1.8382     & 0.6591     \\
    4H2L      & 32        & 0.7863     & 0.3914     & 1.5701     & 0.5639     \\
              & 16        & 0.8762     & 0.4208     & 2.1198     & 0.7656     \\
\midrule 
              & 128       & 0.6581     & 0.3453     & 1.3435     & 0.4797     \\
              & 64        & 0.6397     & 0.3389     & 1.1461     &0.4286      \\
    4H1L      & 32        & 0.6304     & 0.3399     & 1.2180     & 0.4578     \\
              & 16        & 0.6323     & 0.3347     & 1.1953     & 0.4483     \\
\bottomrule
\end{tabularx}
\end{table*}

\renewcommand{\arraystretch}{0.8} 
\begin{table*}[ht] 
\centering
\scriptsize 
\caption{Performance comparison for different input dropout rates with fixed model dropout (0.3). Architectures and embedding dimensions varied.}
\label{tab:input_dropout_templated}
\begin{tabularx}{1.2\columnwidth}{cc|cccc} 
\toprule
\textbf{Architecture} & \textbf{$d_{emb}$} & \textbf{avgADE} $(\downarrow)$ & \textbf{minADE} $(\downarrow)$ & \textbf{avgFDE} $(\downarrow)$ & \textbf{minFDE} $(\downarrow)$ \\ 
\midrule
\multicolumn{6}{@{}l}{\textbf{Input Dropout = 0.001} (Model Dropout = 0.3 fixed)} \\
\addlinespace[0.8ex]
              & 128       & 0.6024     & 0.3192     & 1.2057     & 0.4640     \\
              & 64        & \textbf{0.5626}     & \textbf{0.3076}     & 1.1507     & 0.4437     \\
    8H4L      & 32        & 0.9883     & 0.4007     & 2.1612     & 0.6919     \\
              & 16        & 0.6427     & 0.3294     & 1.2259     & 0.4650     \\
\midrule 
              & 128       & 0.5845     & 0.3177     & 1.0772     & 0.4087     \\
              & 64        & 0.6030     & 0.3226     & 1.1301     & 0.4327     \\
    4H2L      & 32        & 0.6444     & 0.3369     & 1.0663     & 0.4095     \\
              & 16        & 0.6122     & 0.3313     & 1.0184     & 0.3896     \\
\midrule 
              & 128       & 0.6710     & 0.3438     & 1.1590     & 0.4324     \\
              & 64        & 0.6079     & 0.3282     & 1.0949     & 0.4135     \\
    4H1L      & 32        & 0.6566     & 0.3390     & 1.1797     & 0.4503     \\
              & 16        & 0.6114     & 0.3355     & \textbf{0.9827}     & \textbf{0.3815}     \\
\midrule[\heavyrulewidth] 
\multicolumn{6}{@{}l}{\textbf{Input Dropout = 0.5} (Model Dropout = 0.3 fixed)} \\
\addlinespace[0.8ex]
              & 128       & 0.9177     & 0.4231     & 2.4695     & 0.8926     \\
              & 64        & 0.7440     & 0.3625     & 2.0129     & 0.7494     \\
    8H4L      & 32        & 1.0185     & 0.4542     & 2.6004     & 0.9106     \\
              & 16        & 0.7183     & 0.3590     & 1.9118     & 0.7074     \\
\midrule 
              & 128       & 0.6065     & 0.3235     & 1.3345     & 0.5031     \\
              & 64        & 0.8177     & 0.3954     & 2.2104     & 0.8075     \\
    4H2L      & 32        & 0.5979     & 0.3199     & 1.2657     & 0.4901     \\
              & 16        & 0.8970     & 0.4182     & 2.4617     & 0.8894     \\
\midrule 
              & 128       & 0.6333     & 0.3437     & 1.3264     & 0.4792     \\
              & 64        & 0.6129     & 0.3328     & 1.3216     & 0.5078     \\
    4H1L      & 32        & 0.6005     & 0.3237     & 1.2607     & 0.4893     \\
              & 16        & 0.5858     & 0.3172     & 1.2858     & 0.4959     \\
\bottomrule
\end{tabularx}
\end{table*}

\section*{ACKNOWLEDGMENT}
The research leading to these results was funded by the German Federal Ministry for Economic Affairs and Climate Action and was partially conducted in the project “NXT-AIM”. Responsibility for the information and views set out in this publication lies entirely with the authors.


\bibliographystyle{IEEEtran}
\bibliography{references-sync, references-local}

\end{document}